\documentclass[conference]{IEEEtran}
\IEEEoverridecommandlockouts

\usepackage{cite}
\usepackage{amsmath,amssymb,amsfonts}
\usepackage{algorithmic}
\usepackage{graphicx}
\usepackage{textcomp}
\def\BibTeX{{\rm B\kern-.05em{\sc i\kern-.025em b}\kern-.08em
		T\kern-.1667em\lower.7ex\hbox{E}\kern-.125emX}}

\usepackage{multirow}

\usepackage{listings}
\usepackage{color}

\usepackage[table,xcdraw]{xcolor}
\usepackage[normalem]{ulem}
\useunder{\uline}{\ul}{}

\definecolor{dkgreen}{rgb}{0,0.6,0}
\definecolor{gray}{rgb}{0.5,0.5,0.5}
\definecolor{mauve}{rgb}{0.58,0,0.82}

\begin{document}

\title{An IoT Analytics Embodied Agent Model \\based on Context-Aware Machine Learning}

\author{\IEEEauthorblockN{1\textsuperscript{st} Nathalia Nascimento}
	\IEEEauthorblockA{\textit{Laboratory of Software Engineering (LES)} \\
		\textit{Pontifical Catholic University of Rio de Janeiro (PUC-Rio)}\\
		Rio de Janeiro, Brazil \\
		nnascimento@inf.puc-rio.br}
	\and
	\IEEEauthorblockN{2\textsuperscript{nd} Paulo Alencar}
	\IEEEauthorblockA{\textit{David R. Cheriton School of Computer Science} \\
		\textit{University of Waterloo (UW)}\\
		Waterloo, Canada \\
		palencar@csg.uwaterloo.ca}
	\and
	\IEEEauthorblockN{3\textsuperscript{rd} Carlos Lucena}
	\IEEEauthorblockA{\textit{Laboratory of Software Engineering (LES)} \\
		\textit{Pontifical Catholic University of Rio de Janeiro (PUC-Rio)}\\
		Rio de Janeiro, Brazil \\
		lucena@inf.puc-rio.br}
	\and
	\IEEEauthorblockN{4\textsuperscript{th} Donald Cowan}
	\IEEEauthorblockA{\textit{David R. Cheriton School of Computer Science} \\
		\textit{University of Waterloo (UW)}\\
		Waterloo, Canada \\
		dcowan@csg.uwaterloo.ca}
}

\maketitle

\begin{abstract}
Agent-based Internet of Things (IoT) applications have recently emerged as applications that can involve sensors, wireless devices, machines and software that can exchange data and be accessed remotely. Such applications have been proposed in several domains including health care, smart cities and agriculture. However, despite their increased adoption, deploying these applications in specific settings has been very challenging because of the complex static and dynamic variability of the physical devices such as sensors and actuators, the software application behavior and the environment in which the application is embedded. In this paper, we propose a modeling approach for IoT analytics based on learning embodied agents (i.e. situated agents). 
The approach involves: (i) a variability model of IoT embodied agents; (ii) feedback evaluative machine learning; and (iii) reconfiguration of a group of agents in accordance with environmental context. The proposed approach advances the state of the art in that it facilitates the development of Agent-based IoT applications by explicitly capturing their complex and dynamic variabilities and supporting their self-configuration based on an context-aware and machine learning-based approach.   

\end{abstract}


%
\IEEEpeerreviewmaketitle

\section{Introduction}\label{sec:introduction}

Based on the Google Trends tool, 
the Internet of Things (IoT) \cite{atzori2012social} is emerging as a topic that is highly related to robotics and machine learning. In fact, the use of learning agents has been proposed as an appropriate approach to modeling IoT applications \cite{do2017fiot}. These types of applications address the problems of distributed control of devices that must work together to accomplish tasks \cite{atzori2012social}. This has caused agent-based IoT applications to be considered for several domains, such    as health care, smart cities, and agriculture. For example, in a smart city, software agents can autonomously operate traffic lights \cite{do2017fiot}, driverless vehicles \cite{herrero2008decentralized} and street lights \cite{do2017engineering}.

Agents that can interact with other agents or the environment in which the applications are embedded are called {\itshape embodied agents} \cite{marocco2007emergence,Nolfi2016}. The first step in creating an embodied agent is to design its interaction with an application's sensors and actuators, that is, the signals that the agent will send and receive \cite{Nolfi2016}. As a second step, the software engineer provides this agent with a behavior specification compatible with its body and with the task to be accomplished. However, to specify completely the behaviors of a physical system at design-time and to identify and foster characteristics that lead to beneficial collective behavior is difficult. 
To mitigate these problems, many approaches \cite{marocco2007emergence,oliveira2014symbol,do2017engineering} have proposed the use of evolving neural networks to enable an embodied agent to learn to adapt their behavior based on the dynamics of the environment.

The ability of a software system to be configured for different contexts and scenarios is called {\itshape variability} \cite{galster2014variability}. According to Galster et al. \cite{galster2014variability}, achieving variability in software systems requires software engineers to adopt suitable methods and tools for representing, managing and reasoning about change. However, the number and complexity of variation points \cite{pohl2005software}  that must be considered while modeling agents for IoT-based systems is quite high \cite{ayala2015software}. Thus, ``current and traditional agent development processes lack the necessary mechanisms to tackle specific management of components between different applications of the IoT, bearing in mind the inherent variability of these systems" \cite{ayala2015software}.

In this paper, we propose a self-configurable IoT agent approach in which a machine learning procedure assists a software developer in developing embodied agents for the Internet of Things. The approach involves: (i) a variability model for IoT agents; (ii) feedback-evaluative machine-learning; and (iii) reconfiguration of a group of agents in accordance with environmental context.

We provide more details about our proposed approach in Section III. To evaluate the proposed approach, we present an illustrative example in Section IV.  This section presents the experimental setup, results, and evaluation. The remainder of  this paper is organized as follows. Section II describes the background for the proposed approach. Section V presents the related work. The paper ends with our final remarks in Section VI.

\section{Background} \label{sec:background}
\subsection{Embodied Agents}


Embodied agents have a body and are physically
situated, that is, they are physical agents interacting not only
among themselves but also with the physical environment.
They can communicate among themselves and also with human users. Robots, wireless devices and ubiquitous computing are examples of embodied agents \cite{ecagents}. 

\begin{figure}[!htb]
	\centering
	\includegraphics[width=7.5cm]{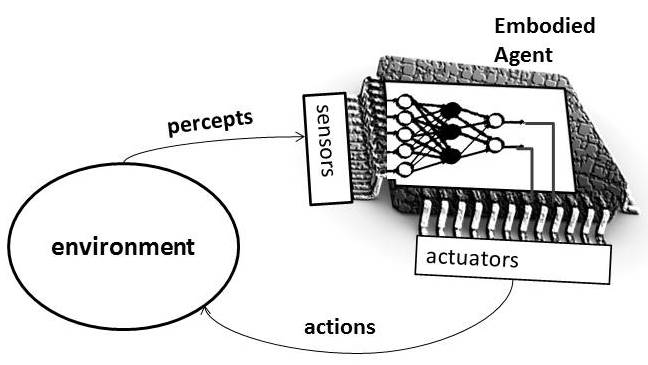}
	\caption{Embodied agent model.}
	\label{figure:embodied}
\end{figure}

Figure \ref{figure:embodied} depicts an embodied agent according to the description presented by the Laboratory of Artificial Life and Robotics \cite{laralsite} about embodied agents. They define embodied agents as agents that have a body and are controlled by an analysis architecture, such as artificial neural networks. These agents use learning techniques, such as an evolutionary algorithm, to adapt to execute a specific task.


\subsection{IoT Embodied Agents}
\begin{figure}[!htb]
	\centering
	\includegraphics[width=6.7cm]{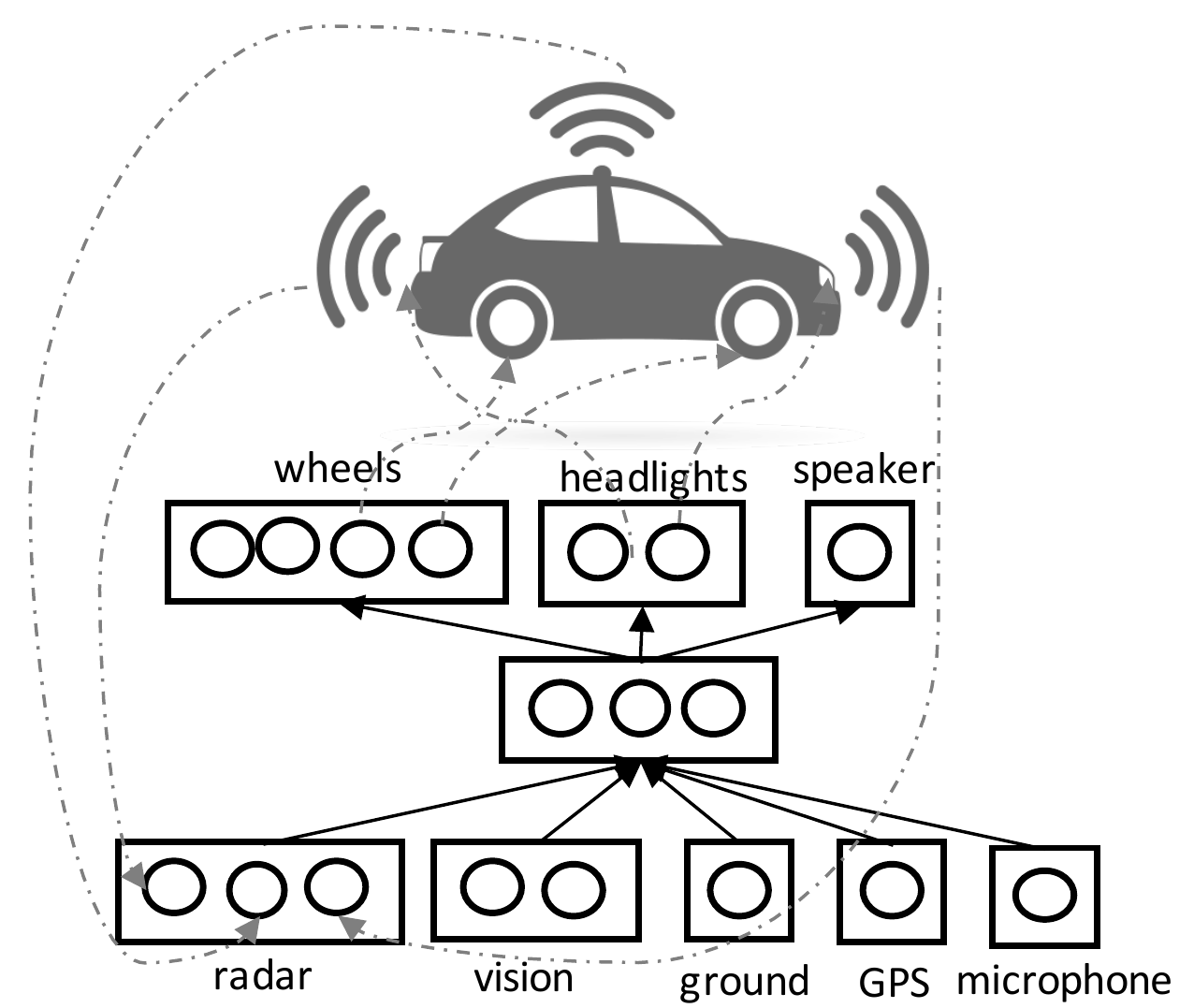}
	\caption{An example of an IoT embodied agent. }
	\label{figure:smartthing}
\end{figure}

According to the description about embodied agents provided in this section, Figure \ref{figure:smartthing} illustrates an IoT embodied agent in a scenario of autonomous cars. In this example, the body of the agent is a car with four wheels, GPS, headlights, etc. As described above, an embodied agent must have a local analysis architecture to sense the environment and behave accordingly. In such example, the autonomous car uses an artificial neural network. There is an input neuron for each one of the car's sensors and an output neuron for each one of the motors and actuators. The neuron output values may determine the direction of the wheels and whether the car turn on the headlights. 

\subsection{Context-aware Approaches}
Context is any information that can be used to characterize the state of different entities, such as persons, places or physical objects \cite{abowd1999towards}. Context-aware computing is used to represent a system that understands the context and takes an action based on that particular context \cite{sezer2018context}.

Nascimento et al. \cite{nascimento2018context} investigate the advantages of considering the context in a machine learning-based approach. Accordingly, a system that is able to reconfigure its analysis model according to the context outperforms a system that uses an unique and versatile model.

\subsection{Reconfigurable Systems}
To achieve variability in a system, the first step is to understand and represent variability in its application domain. Our approach incorporates feature-oriented domain analysis (FODA) \cite{pohl2005software} to represent the software's variability. According to the FODA notation, features can be classified as mandatory, optional and alternative. Alternative features are not to be used in the same instance. 

\section{Approach} \label{sec:approach}

We aim to support the development of embodied agents to work in real or simulated scenarios related to the Internet of Things. Therefore, we designed a platform to support  i) handling variability in IoT embodied agents, ii) selecting the physical components that will compose each agent, and iii) finding their appropriate behavior according to their bodies and the scenario where they will be applied. 


\subsection{Variability in IoT Embodied Agents}

In an Internet of Things application suite, there are several options for physical components and software behaviors for the design of a physical agent \cite{ayala2015software}. According to existing experiments \cite{soni2017smart} and our experience with the IoT domain \cite{do2017fiot,nathalia2016iot,do2017engineering}, we introduce possible variants of an IoT embodied agent and we use a FODA notation to explore this variability. 

\begin{figure*}[!htb]
	\centering
	\includegraphics[width=14.5cm]{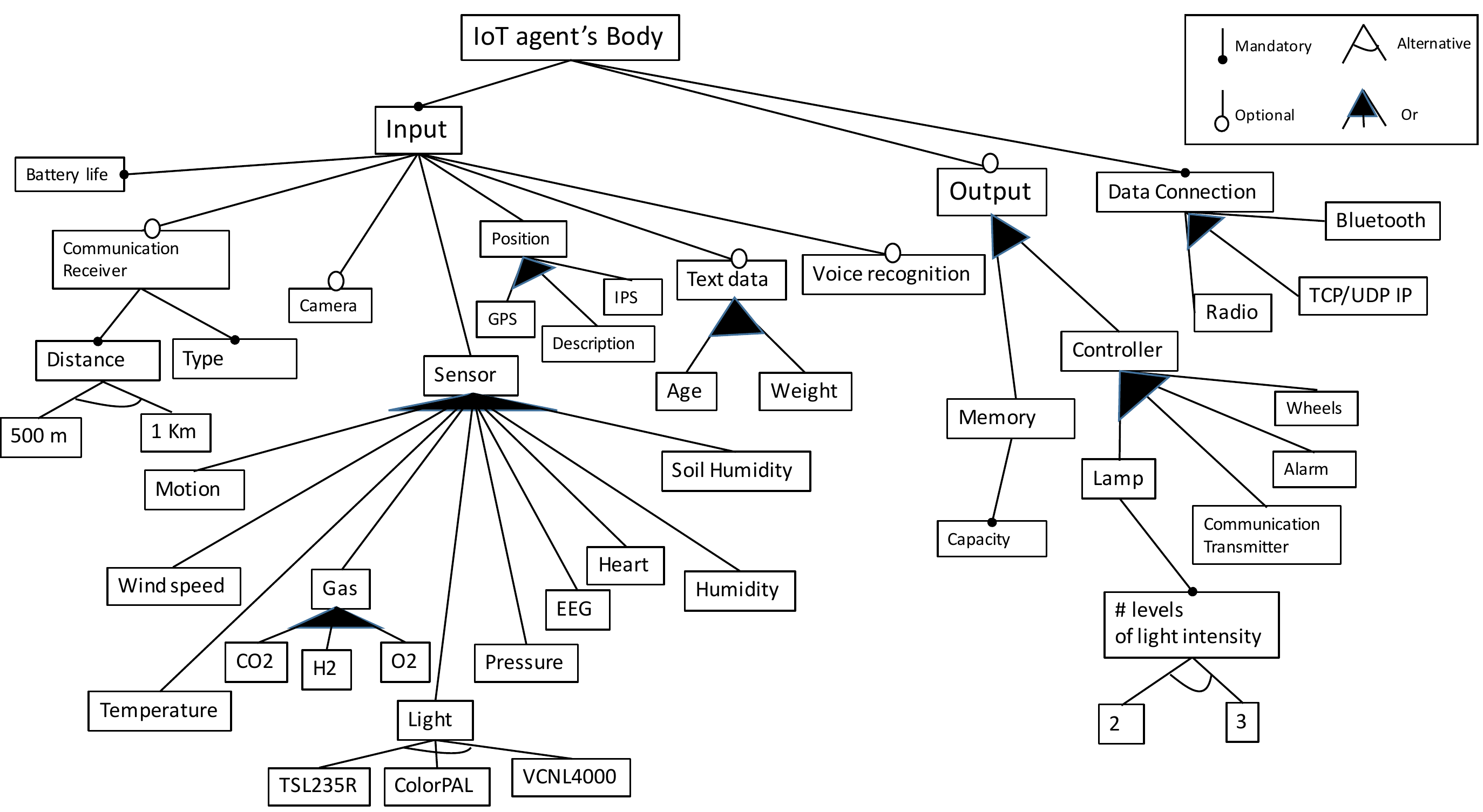}
	\caption{Feature model of an IoT embodied agent's body. }
	\label{figure:bodyset}
\end{figure*}	

\begin{figure*}[!htb]
	\centering
	\includegraphics[width=14.5cm]{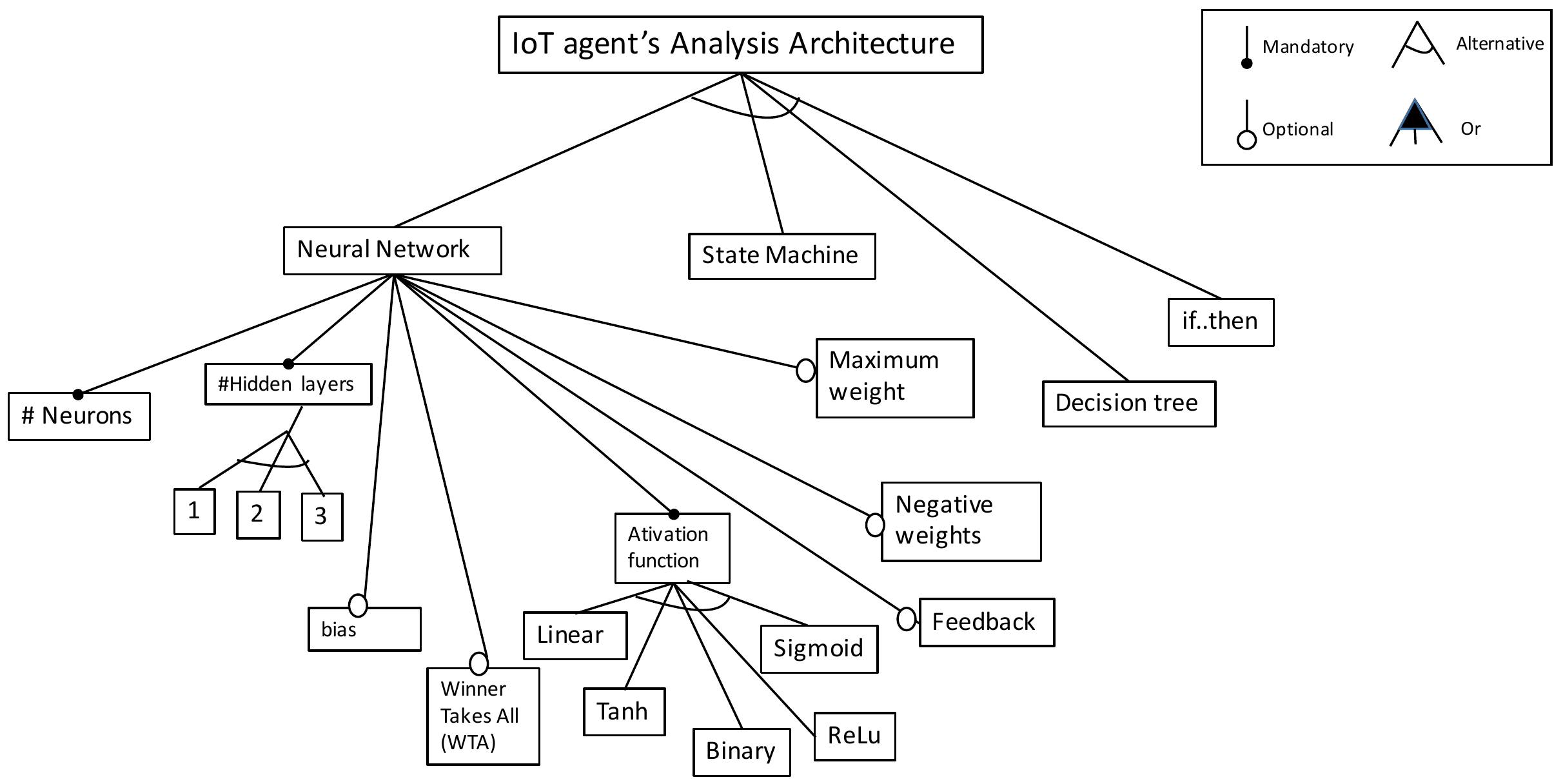}
	\caption{Feature model of an IoT embodied agent's analysis architecture. }
	\label{figure:analysisset}
\end{figure*}

The physical devices may vary in terms of the types of sensors, such as temperature and humidity, and in terms of actuators, as depicted in the feature diagram presented in Figure \ref{figure:bodyset}  . Each sensor can also vary in terms of brands, changing such parameters as  energy consumption and battery life. 
Note that, depending on the application domain, this feature diagram may contain different and more specific features. For example, to  create smart street light agents, we can provide a different version of this feature model, discarding some options of sensors, such as heart, EEG and pressure sensors.



In addition, we also need to deal with variants in agent architecture that the agent uses to sense the environment and  behave accordingly. For example, this architecture can be a decision tree, a state machine or a neural network. Many approaches \cite{marocco2007emergence,do2017engineering} use {\itshape neuroevolution}, which is ``a learning algorithm which uses genetic 
algorithms to train neural networks" \cite{whiteson2005evolving}). This type of network determines the behavior of an agent automatically based on its physical characteristics and the environment being monitored. A neural network is a well-known approach to provide responses dynamically and automatically, and create a mapping of input-output relations, which may compactly represent a set of ``if..then" conditions \cite{do2017engineering}, such as: ``if the temperature is below 10$^{\circ}$C, then turn on the heat." However, finding an appropriate neural network architecture based on the physical features that were selected for an agent, is not easy. To model the neural network, we also need to account for its architectural variability, such as the activation function, the number of layers and neurons and properties such as the use of winner-take-all (WTA) as a neural selection mechanisms \cite{fukai1997simple} and the inclusion of recurrent connections \cite{marocco2007emergence}. We explore these variabilities in Figure \ref{figure:analysisset}.

With respect to analysis variabilities, Marocco and Nolfi \cite{marocco2007emergence}, performed two experiments with the same embodied agents, varying only the neural network architectures and neural activation functions. In the first experiment, they used a neural network without internal neurons, while in the second experiment, they used a neural network with internal neurons and recurrent connections. In addition, they also used different functions to compute the neurons' outputs. Based only on the neural network characteristics, they classified the robots from the first experiment as reactive robots (i.e. ``motor actions can only be determined on the basis of the current sensory state"), and non-reactive robots (i.e. ``motor actions are also influenced by previous sensory and internal states"). Marocco and Nolfi \cite{marocco2007emergence} analyzed whether the type of neural architecture influenced the performance of a team of robots. They showed that the differences in performance between reactive and non-reactive robots vary according to the environmental conditions and how the robots have been evaluated.

Oliveira and Loula \cite{oliveira2014symbol} investigated symbol representations in communication based on the neural architecture topology that is used to control an embodied agent. They found that the communication system varies according to how the hidden layers connect the visual inputs to the auditory inputs.

Jarraya et al. \cite{Jarraya2018} propose a multiagent approach for pervasive computing that aims to identify human activities in smart homes. In their approach, the set of agents must observe sensor data and make local predictions. Jarraya et al. \cite{Jarraya2018} state that ``depending on the nature of sensor data, agents may hold different types of classifiers."

These findings have helped us to conclude that to support the design of IoT embodied agents, we need to account for the variability of the physical body and the architecture that analyses the inputs. In addition, we also concluded that the analysis architecture cannot be considered as a black box in the system, since its structure must fit the characteristics that were selected to compose the body and behavior of the agent. 

\subsection{Architecture} \label{sec:architecture}

\begin{figure*}[!htb]
	\centering
	\includegraphics[width=16.2cm]{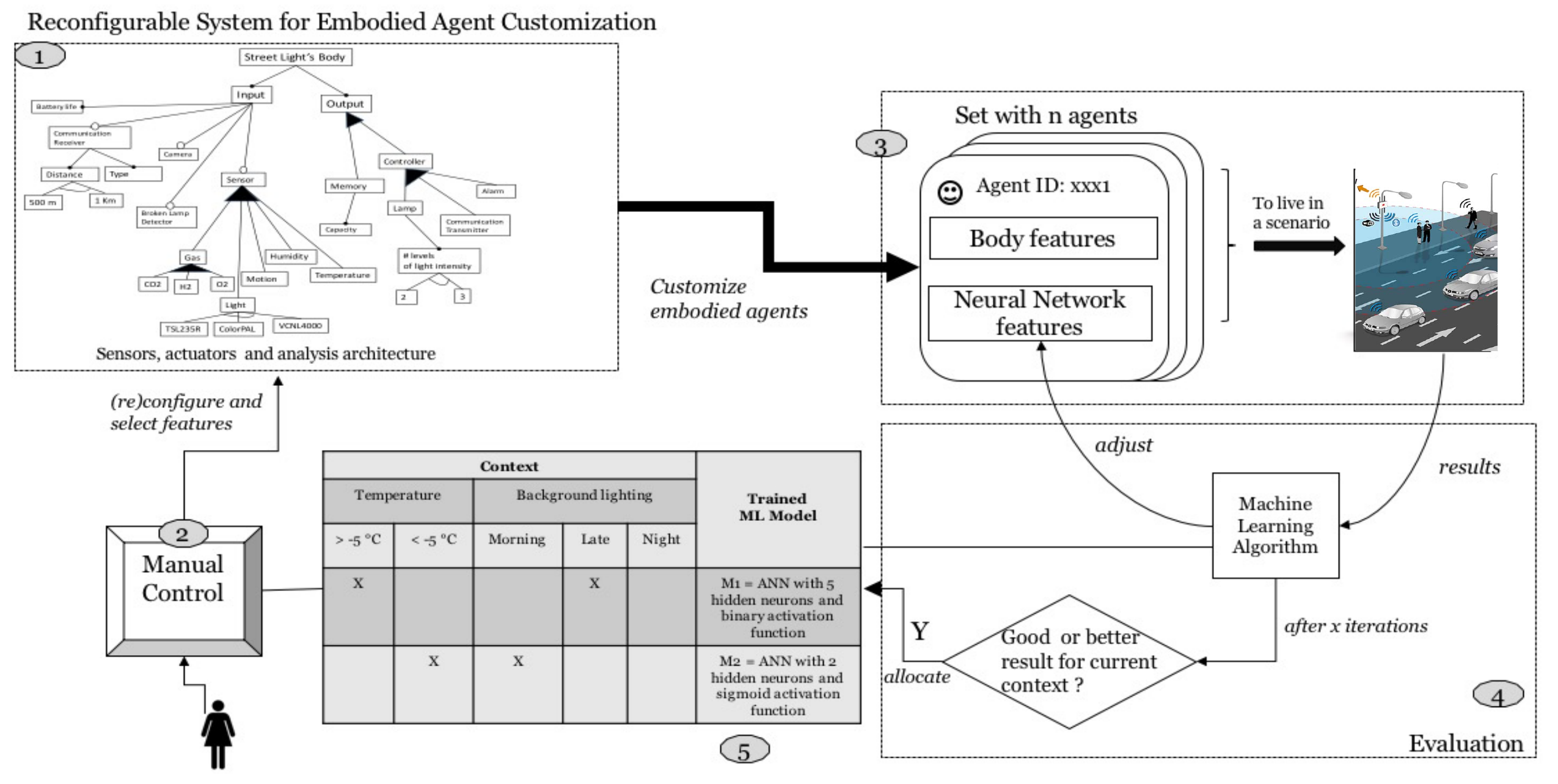} 
	\caption{High-level model of the self-configurable agent approach to generate embodied agents. }
	\label{figure:approach}
\end{figure*}

Based on the two main variation points we have identified, we propose a platform to design embodied agents applications based on the Internet of Things. 
Figure \ref{figure:approach} depicts the high-level model of our proposed approach. 
Basically, this platform contains five modules: i) a reconfigurable system that contains the characteristics that can be used to compose the set of agents, according to the application domain; ii) a manual control that allows an IoT expert to select the first set of features manually;  iii) the creation of a set of agents containing the selected characteristics that are also able to use a neural network to learn about the environment; iv) a module for evaluating feedback tasks, by investigating the performance of the group of agents in the application scenario during the learning execution. The evaluation process has to be implemented according to the application and the learning algorithm. For example, if an application for automobile traffic control has the goal of reducing urban traffic congestion, the evaluation may be performed based on the number of vehicles that had finished their routes in a specific period; and (v) a module to store and retrieve machine learning models based on the context, as described in \cite{nascimento2018context}. It allows the set of agents to switch the analysis architecture according to the context at runtime. The IoT expert can also use this context information to reconfigure the set of agents.

\subsection{A Context-Aware Machine Learning Embodied Agent Model}
\begin{figure}[!htb]
	\centering
	\includegraphics[width=9.0cm]{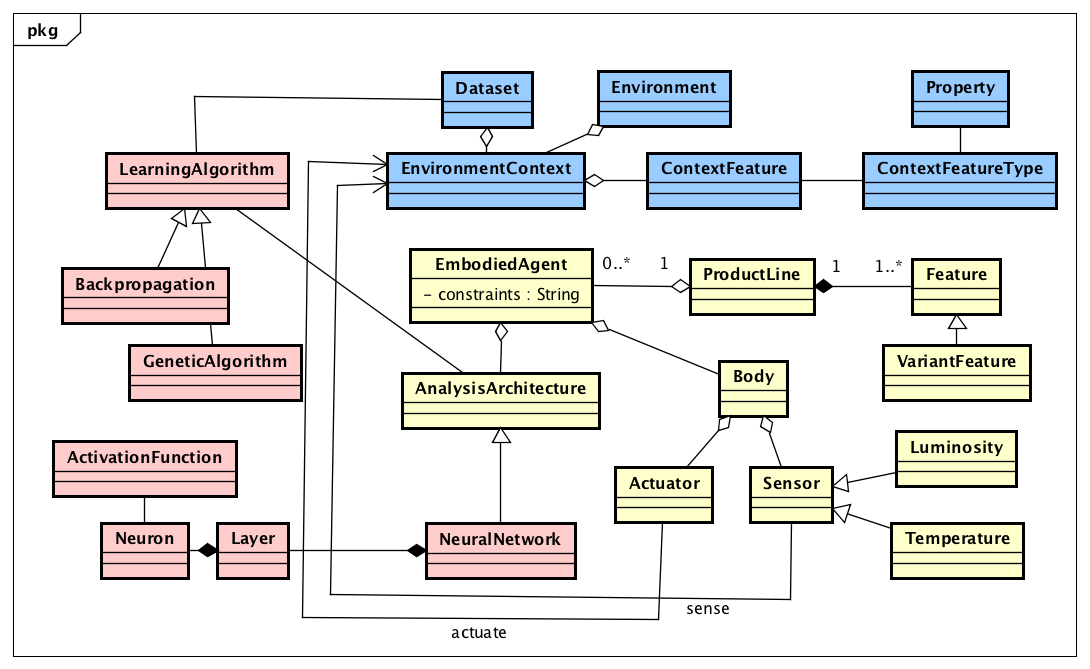}
	\caption{Class diagram of the context-aware machine learning embodied agent for IoT analytics.}
	\label{figure:classdiagram}
\end{figure}

Figure \ref{figure:classdiagram} illustrates the class diagram of the proposed model. We separated the classes into three modules: 
 (i) a reconfiguration system, that contains the set of characteristics to compose each embodied agent; (ii) the environment, that is the union of a set of contexts; and (iii) the machine learning module, that is associated with the environment and reconfiguration system classes. 
 Accordingly, the product line has some fixed and variant features that can be used to compose the embodied agents. An embodied agent has sensors and actuators to sense and act on the current context. While the agent interacts with the environment, the learning algorithm can adjust the agent's analysis architecture in order to improve its performance to the current context. 

\section{Illustrative Example: Smart Street Lights}

We selected one example from the IoT domain: a smart street light application. Our goal is to show how this application will be executed by using our proposed model. 
In this application scenario, we consider a set of street lights distributed in a neighborhood. For more details concerning
this application scenario, see \cite{do2017engineering}.

Each street light represents an IoT agent, which needs to operate in an environment composed of different contexts. For example, sometimes the background light can be bright and at other times dark. With respect to the environment background light, the application scenario has some variants: (i) night (background light is equal to 0.0); (ii) late afternoon (background light is equal to 0.5); and (iii) morning (background light is equal to 1.0). Each street light contains a lighting sensor, but its local brightness also interferes on the sensor measurement.

\subsection{Selecting Physical and Neural Network Features}

An IoT expert selected three physical inputs and two physical outputs to measure and operate each one of the street lights. In addition, the engineer selected a neural network with one hidden layer with ten units as the initial network for each agent with the sigmoid function as the activation function of this neural network.

\subsection{Learning about  the environment}
During the training process, the algorithm evaluates the options for weights of the network based on energy consumption, the number of people that finished their routes before the simulation ends, and the total time spent by people moving during their trip. 

\subsubsection{Adjusting the agents to an initial context}
During a first simulation, while the background light was always bright, the collection of street lights found a solution that provided a performance, say X+1, during the morning.
\subsubsection{Environmental context changes}
\begin{figure}[!htb]
	\centering
	\includegraphics[width=8.6cm]{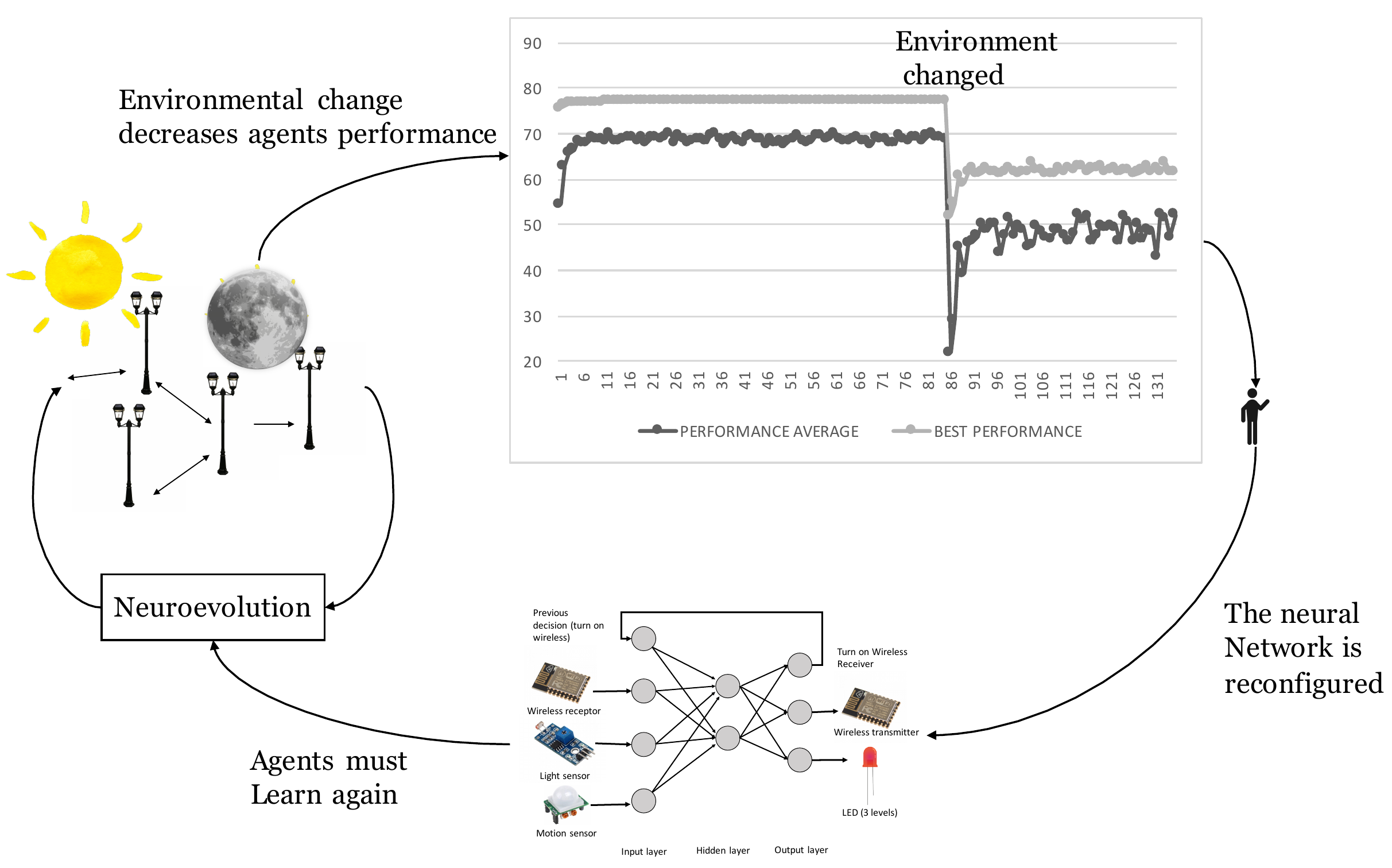}
	\caption{Reconfiguring the set of features.}
	\label{figure:secondlearninginteraction}
\end{figure}

After a time, a change in the environment occurred. Now, these agents are operating in an environment in which sometimes the background light can be bright and at other times dark. 
After the environment changed to the night, the lights' solution was adjusted to deal with this change. However, this new generic solution decreased the performance, say X, during the morning, as shown in Figure \ref{figure:secondlearninginteraction}. 


In a traditional learning approach, the street light would be unable to return to its previous configuration, as the street light would not maintain different versions of its configuration. However, 
 the configuration history that is supported by our approach could enable the street light to switch its analysis architecture to specialized solutions that were trained for each one of the background lighting variants.
 

\subsection{Reconfiguring the set of agents for the new context}
\begin{figure}[!htb]
	\centering
	\includegraphics[width=11.6cm]{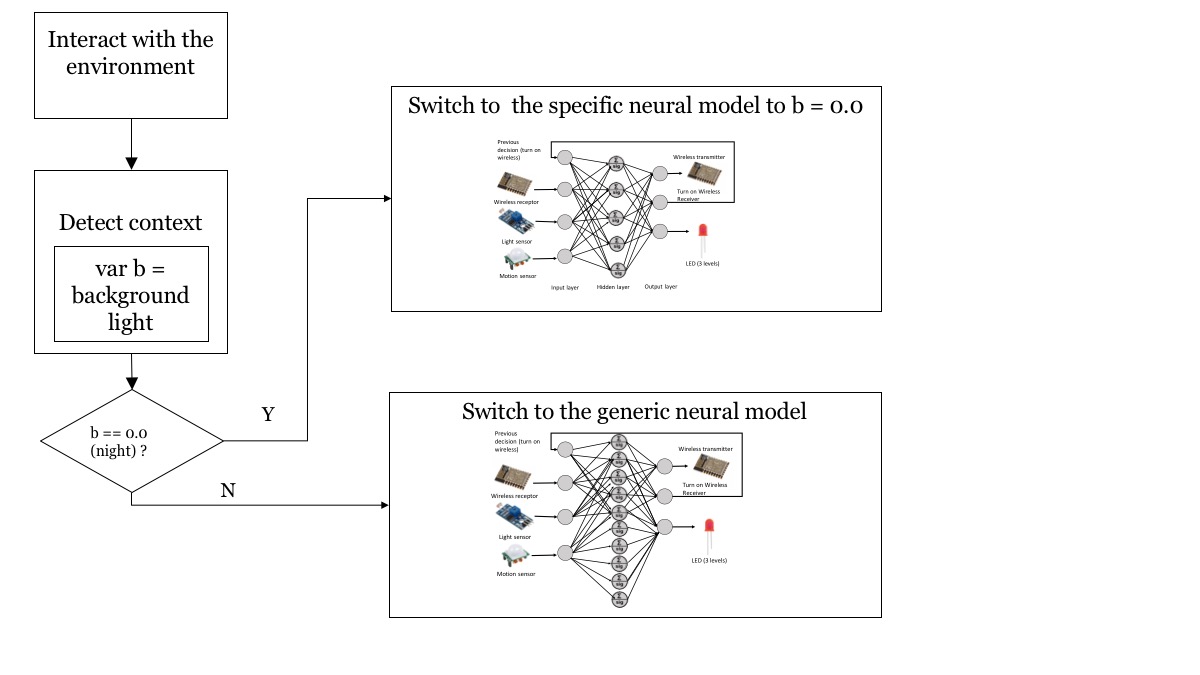}
	\caption{Selecting the analysis architecture according to the context.}
	\label{figure:changeContext}
\end{figure}

As our approach enables the set of agent to use more than one analysis model at runtime, 
the expert can provide a new model to be trained for this new context. For instance, he maintained the number of sensor inputs, but selected different variants for the neural network, such as the number of neurons in the hidden layer.
Then, the learning algorithm was re-executed and the agents were able to select a new model to cope with this environmental change, as depicted in Figure \ref{figure:changeContext}.

\section{Contributions and Ongoing Work}
To handling variability in IoT embodied agents, we identified the main variation points of these kinds of applications, including the variants that can be involved in a neural network design. We also provided a feature-oriented variability model, which is an established software engineering module.

In addition, we proposed an approach that takes context into account to train and deploy machine learning-based models for IoT embodied agents. To demonstrate the use of this context-aware approach, we reproduced an experiment showing how this application operates by taking the main steps of our proposed approach into account. 




Our next step is to select more state-of-the-art experiments and verify if our proposed approach can improve their results. In addition, we also want to consider to enable the use of a learning technique to reconfigure the set of features based on environmental changes automatically. As we proposed a hybrid architecture, we can use this learning technique only to reconfigure the variants related to one of the variation points, such as the neural network properties. In such an instance, we can have an expert responsible for handling the body and behavior variability of the IoT agents.

\section*{Acknowledgment}

This work has been supported by the Laboratory of Software Engineering (LES) at PUC-Rio. It has been developed in cooperation with the University of Waterloo, Canada. Our thanks to CNPq, CAPES, FAPERJ and PUC-Rio for their support through scholarships and fellowships.

\bibliographystyle{IEEEtran}
\bibliography{sigproc}

\end{document}